\documentclass[11pt,a4paper]{article}
\usepackage[hyperref]{acl2019}
\usepackage{times}
\usepackage{latexsym}
\usepackage{graphicx}  %
\usepackage{amsmath}
\usepackage{amssymb}
\usepackage{bbm}
\usepackage{color}
\usepackage{subcaption}
\usepackage{xspace}
\usepackage{url}
\usepackage{booktabs}

\aclfinalcopy %
\def\aclpaperid{2158} %

\title{Finding Your Voice:\\
 The Linguistic Development of Mental Health Counselors}

\author{Justine Zhang \\
  Cornell University \\
  \small{\tt jz727@cornell.edu} \\\And
  Robert Filbin \\
  Crisis  Text Line \\
  \small{\tt bob@crisistextline.org} \\\And
  Christine Morrison \\
  Crisis Text Line \\
  \small{\tt christine@crisistextline.org} \\\AND
  Jaclyn Weiser \\
  Crisis Text Line \\
  \small{\tt jweiser@crisistextline.org} \\\And
  Cristian Danescu-Niculescu-Mizil \thanks{\ \ Corresponding senior author.}\\
  Cornell University \\
  \small{\tt cristian@cs.cornell.edu}  \\}

\date{}

\begin{document}

\definecolor{marsala}{RGB}{150, 79, 76}
\definecolor{navy}{RGB}{0, 0, 128}
\definecolor{turquoise}{RGB}{69, 181, 170}
\definecolor{forest}{RGB}{25, 150, 25}

\newcommand{\cut}[1]{}

\newif\ifshowchanges
\showchangesfalse
\newcommand{\cameraready}[1]{{\textcolor{turquoise}{#1}}}
\newcommand{\camerareadynew}[1]{{\textcolor{forest}{#1}}}
\ifshowchanges
\else
\renewcommand{\cameraready}[1]{{\textcolor{black}{#1}}}
\renewcommand{\camerareadynew}[1]{{\textcolor{black}{#1}}}
\fi

\newif\ifshowcomments
\showcommentstrue
\newcommand{\todo}[1]{\xspace\textcolor{red}{#1}}
\newcommand{\cd}[1]{{\textcolor{blue}{#1}}}
\newcommand{\justine}[1]{{\textcolor{marsala}{#1}}}

\newcommand{\todofig}{\xspace\textcolor{red}{Figure F}\xspace}
\newcommand{\todotab}{\xspace\textcolor{red}{Table T}\xspace}
\newcommand{\todostat}{\xspace\textcolor{red}{\#}\xspace}
\newcommand{\todoprop}{\xspace\textcolor{red}{\%}\xspace}
\newcommand{\todoparam}{\xspace\textcolor{red}{\$}\xspace}
\newcommand{\todocite}[1]{\xspace\textcolor{red}{CITE(#1)}\xspace}
\newcommand{\todoegs}{\xspace\textcolor{red}{example}\xspace}
\newcommand{\wording}[2]{\xspace\textcolor{navy}{[#1$\mid$#2]}}
\newcommand{\question}[1]{\xspace\textcolor{turquoise}{\{#1\}}}
\ifshowcomments
\else
\renewcommand\todo[1]{}
\renewcommand{\cd}[1]{}
\renewcommand{\justine}[1]{}
\renewcommand{\todofig}{}
\renewcommand{\todotab}{}
\renewcommand{\todostat}{}
\renewcommand{\todoprop}{}
\renewcommand{\todoparam}{}
\renewcommand{\todocite}[1]{}
\renewcommand{\todoegs}{}
\renewcommand{\wording}[2]{#1}
\renewcommand{\question}[1]{}
\fi

\newcommand{\xhdr}[1]{{\noindent\bfseries #1.}}
\newcommand{\xhdrq}[1]{{\noindent\bfseries #1?}}

\newcommand{\overbar}[1]{\mkern 1.5mu\overline{\mkern-1.5mu#1\mkern-1.5mu}\mkern 1.5mu}

\newcommand{\inter}[1]{\ensuremath{\mathcal{#1}}\xspace}
\newcommand{\LM}[2]{\ensuremath{\mathcal{L}_\inter{#1}^{#2}}\xspace}
\newcommand{\crossent}[2]{\ensuremath{H(#1,#2)}\xspace}
\newcommand{\lifestage}[1]{\ensuremath{\mathbb{S}_{#1}}\xspace}
\newcommand{\ceillifestage}{\ensuremath{\mathbb{S}_{\max}}\xspace}
\newcommand{\maxlifestage}{\ensuremath{\overbar{\mathbb{S}}}\xspace}
\newcommand{\prup}[1]{\ensuremath{\uparrow_{#1}}\xspace}

\newcommand{\wordfreq}[2]{\ensuremath{\Pr(\egword{#1},#2)}\xspace}
\newcommand{\egword}[1]{\textit{#1}}
\newcommand{\egwordnew}[1]{\textcolor{red}{\textit{#1}}}
\newcommand{\egwordold}[1]{\textcolor{blue}{\textit{#1}}}

\newcommand{\vocab}{\ensuremath{\mathbb{V}}\xspace}
\newcommand{\segvocab}[1]{\ensuremath{\mathbb{V}_{#1}}\xspace}

\newcommand{\firstseg}{\textcolor{black}{\small\texttt{hello}}\xspace}
\newcommand{\secondseg}{\textcolor{black}{\small\texttt{problem exploration}}\xspace}
\newcommand{\thirdseg}{\textcolor{black}{\small\texttt{goal identification}}\xspace}
\newcommand{\fourthseg}{\textcolor{black}{\small\texttt{problem solving}}\xspace}
\newcommand{\fifthseg}{\textcolor{black}{\small\texttt{goodbye}}\xspace}

\maketitle
\begin{abstract}
Mental health counseling is an enterprise with profound societal importance where conversations play a primary role.  
In order to acquire the conversational skills needed to face a challenging range of situations, mental health counselors must rely on training and on continued experience with actual clients.   
However, in the absence of large scale longitudinal studies, the nature and significance of this developmental process remain unclear.  
For example,  prior literature suggests  that experience might not translate into consequential changes in counselor behavior.  
This has led some to even argue that counseling is a profession without expertise. 

In this work, we develop a computational framework to quantify the extent to which individuals change their linguistic behavior with experience and to study the nature of this evolution. 
 We use our framework to conduct a large longitudinal study of mental health counseling conversations, tracking over 3,400 counselors across their tenure.  
 We reveal that overall, counselors do indeed change their conversational behavior to become more diverse across interactions, {developing} an individual voice that distinguishes them from other counselors.
 Furthermore, a finer-grained investigation shows that the rate and nature of this diversification vary across functionally different conversational components.

\end{abstract}

\section{Introduction}
\label{sec:intro}

Conversations are central to many human endeavors and professions, from academic advising, to business negotiations, to customer service, to mental health counseling.  
The choice of conversational language has been shown to correlate with a broad spectrum of
consequential outcomes such as success in persuasion \cite{habernal_what_2016,tan_winning_2016,zhang_conversational_2016,packard_im_2018}, in team performance \cite{yang_weakly_2015,niculae_conversational_2016}, or in social support \cite{atkins_scaling_2014,althoff_large-scale_2016,choudhury_language_2017}.
However, surprisingly little is known about how adults develop conversational skills.  
Understanding this challenging process has the potential to help with tracking and fostering behavioral development in conversation-focused endeavors.

In this work, we address this knowledge gap by examining how individuals change their conversational language in a domain with profound societal importance, where conversations play a primary role: mental health counseling. {Counselors in this setting} face a daunting task: through conversations, they need to empathetically respond to and support clients undergoing psychological distress \cite{althoff_large-scale_2016}.

Initial training can orient counselors towards the principles of {counseling conversations,}
but cannot cover the broad range of situations they {will} inevitably encounter.
Practical experience, on the other hand, comes with scarce direct feedback: 
it is hard for counselors to gauge whether a client was positively affected by a conversation.
 Perhaps as a result, prior studies find no consequential change in counselor behavior or effectiveness beyond their initial training \cite{dawes_house_1996,hill_is_2015,goldberg_psychotherapists_2016}.   
This lack of evidence has led some to even argue that psychological counseling is a ``profession without any expertise'' \cite{tracey_expertise_2014}. 

However, even in the absence of feedback, multiple forces could lead counselors to change their behavior with experience.   
First, practice exposes---and perhaps familiarizes---the counselor to a wide range of client situations, far beyond what training can encompass. 
{This could enable them to increasingly adapt  to the situations they encounter, differentiating their behavior in one conversation from their behavior in another (\textit{within-counselor diversification}).}

Second, counselors come into the domain with individual personalities and conversational styles.  
A strict adherence to training materials can lead to unnatural and robotic conversations, which can be taxing on the counselors \cite{orlinsky_how_2005} and detrimental to the client \cite{henry_effects_1993,castonguay_predicting_1996}.   
A counselor might therefore seek to adapt training principles to their own personality, finding a voice that distinguishes them from other counselors (\textit{between-counselor~diversification}). 

However, change is far from guaranteed,
and several forces potentially counteract the two vectors of change outlined above. 
The {challenging nature and} high (sometimes literally life-and-death) stakes of the conversations might lead to an overly cautious approach, deterring counselors from departing from language exemplified in the training material and {instead} leading them to seek  ``security in a restrictive practice routine'' \cite{skovholt_searching_1997}.  
The lack of feedback might exacerbate this stagnation and inflexibility, since counselors have no direct way to assess the potential effects of behavioral changes \cite{shanteau_running_1992,kahneman_conditions_2009}. 

In this work, we design a computational methodology to quantitatively track systematic changes along {the two dimensions of} within- and between-counselor linguistic diversification.
Our framework additionally exposes the rate at which these changes happen across individuals and highlights 
particular functional components of their language where this change is especially salient.

We apply this framework to conduct a large-scale longitudinal study of mental health counseling conversations on {Crisis Text Line}, a text-based crisis counseling platform,\footnote{This study was done in collaboration with {Crisis Text Line}, using anonymized data and following IRB approval. {Details of the data access are included in Section \ref{sec:data}.}} tracking the conversational behavior of over 3,400 counselors across their tenure.
Our study reveals that overall, counselors do indeed systematically change their conversational behavior, becoming more linguistically diverse across their own conversations as well as among each other. 
Furthermore, this diversification process advances at different rates in different functional components of the conversation; in particular, within-counselor diversification is amplified in components concerning the client's specific problems. This suggests that counselors accumulate domain fluency, perhaps enabling them to better address the particular situations they encounter. 

{We complement this intuition by examining how counselor vocabulary shifts with experience: which words increase in usage or fall out of favor as counselors gain experience? 
We find that as they advance in tenure, counselors adopt more specialized as well as more colloquial language, while slowly abandoning language from the training material.}

Overall, these results provide the first evidence of a systematic development of mental health counselors with experience, thus contributing to open questions about expertise from the counseling and psychotherapy literature.
From a practical standpoint, our framework could aid counseling platforms to automatically identify counselors that stagnate. 
Additionally, by highlighting conversational practices that require more experience to grasp,  this type of understanding can inform sustained training programs across a counselor's career \cite{tracey_expertise_2014}.

More broadly, our work takes initial steps in understanding how individuals develop their conversational behavior.   Beyond mental health counseling, this methodology can potentially provide  insights into other domains where {having conversations is crucial} but potentially hard to teach, such as academic advising and customer service.

\section{Further Related Work}
\label{sec:related}

Our work draws on prior literature concerning language use in conversational settings like counseling, as well as on studies of linguistic change. 

\xhdr{Language in the mental health domain}
Prior literature has underlined the importance of linguistic choices and conversational behavior in mental health-related contexts. Qualitative \cite{labov_therapeutic_1977,miller_small_2001,catley_adherence_2006,gaume_counselor_2010} and computational \cite{atkins_scaling_2014,tanana_recursive_2015,tanana_comparison_2016,althoff_large-scale_2016,perez-rosas_analyzing_2018} studies of dialogic interactions in counseling and psychotherapy have highlighted the potential benefits of conversational skills---such as a counselor's ability to reflect on a client's particular concerns. Other work has considered a broader range of settings such as online forums, exploring how psychological support is sought and provided \cite{choudhury_language_2017,ernala_linguistic_2017,yates_depression_2017,yang_commitment_2017,yang_seekers_2019,pruksachatkun_moments_2019}. While these studies have focused on linguistic behaviors in the scope of single interactions, we seek to understand how these behaviors develop over time. 
\xhdr{Linguistic behavior in conversations}
Our work relates more broadly to other computational studies of conversations. As in the mental health setting, these studies have largely considered the behavior of interlocutors in a single conversation, tying it to outcomes such as persuasion, problem-solving and incivility \cite{curhan_thin_2007,rosenthal_i_2015,wang_winning_2017,zhang_conversations_2018}, or to social correlates such as power, gender and age \cite{gonzales_language_2010-2,ireland_language_2011,danescu-niculescu-mizil_echoes_2012,nguyen_how_2013,prabhakaran_gender_2014}. In contrast, our work seeks to chart the development of linguistic tendencies of individuals as they engage in many conversations. 

\xhdr{Linguistic change}
Prior studies have examined language change in offline \cite{labov_social_1966,labov_sociolinguistic_1972,labov_principles_2011,eckert_variation_2005,tagliamonte_variationist_2011} and online \cite{cassell_language_2006,lam_language_2008,nguyen_language_2011,garley_beefmoves_2012,danescu-niculescu-mizil_no_2013,bamman_gender_2014,eisenstein_diffusion_2014,nguyen_computational_2015,kulkarni_statistically_2015,kulkarni_freshman_2016,goel_social_2016} settings. A core component of such studies of linguistic \emph{socialization} has been the presence of external social influences that may impose {linguistic norms---for instance, individuals are often able to observe the language used in posts by other members of an online community.}
In contrast, such social pressures are minimized in the counseling domain we consider. Our study hence reflects an alternate set of possible mechanisms for linguistic change, disentangling individual experience from normative influence.

\section{Domain: Mental Health Counseling}
\label{sec:data}

In this work we study the linguistic development of counselors in {Crisis Text Line}, a large text-based mental health  counseling platform. The platform offers a free 24/7 service for individuals undergoing mental health crises---henceforth \textit{texters}---to have one-on-one conversations with counselors via text messages. In collaboration with the platform, and in accordance with an IRB, we accessed the complete collection of anonymized conversations up to September 2018.\footnote{{This study was done as part of a research fellowship program organized by Crisis Text Line, which granted the authors access to an anonymized version of the dataset, subject to an IRB. The program is open to other researchers by application:} \footnotesize{\url{https://www.crisistextline.org/open-data-collaborations}}. \citet{pisani_protecting_2019} {describes the extensive privacy and ethical considerations, and the policies implemented by the platform to address them.}
 }

\xhdr{Counselors}
The service is driven by a dedicated roster of mental health counselors, comprised of volunteers who join the platform via an application process.\footnote{Just over half of counselors have no prior experience in a psychology-related domain. Our results are qualitatively similar over the subset of counselors who do not have a psychology-related background.} Each counselor starts by completing a standardized training curriculum; once they graduate they can then sign up for shifts during which they take conversations with texters. To gain a better understanding of the domain, the first author completed the training curriculum and participated in a shift.

Many counselors are quite committed to the service: the median counselor takes 43 conversations while a quarter take at least 120. In taking a longitudinal view, we focus on this latter subpopulation of counselors who have taken at least 120 conversations, and who started their tenure after July 2015.\footnote{We enforce the start time to account for variations in the training curriculum; our results are qualitatively similar under slight modifications to the start month, and over counselors with shorter tenures.} 
{The 3,475 counselors in this subset took their first 120 conversations over an average of 5 months; in total they have taken 1,055,924  conversations, accounting for 73\% of all the conversations on the platform to date.}
We analyze a counselor's behavior at different experience levels by dividing their tenure into a series of consecutive \textit{life-stages} $\lifestage{0}, \lifestage{1}, \ldots$ of $s$ conversations each; unless otherwise indicated we take $s=20$. 
Our particular focus is on examining counselor behavior over their first 120 conversations. We refer to a counselor as \textit{tenured} once they have taken at least 100 conversations, and denote the life-stage consisting of a counselor's 100th to 120th conversations as \maxlifestage. Our subsequent analyses compare behavior in earlier life-stages to \maxlifestage, using this eventual tenured state as a reference point.

\xhdr{Counseling conversations}
In each conversation, a counselor is faced with a challenging task: guiding a texter through a moment of crisis towards a better mental state. 
This process requires the counselor to empathetically engage with the texter, often in difficult, high-stakes situations. The complexity of this inherently interactional task gives rise to substantive conversations---averaging 14 counselor messages per conversation  and 28 words per counselor message---that yield a rich array of linguistic behaviors \cite{atkins_scaling_2014,althoff_large-scale_2016}. We focus on tracking counselor behavior over conversations with at least 10 counselor messages (comprising 64\% of our dataset).

The challenge of taking conversations is compounded by the diversity of texters who contact the service. Texters can come with a wide range of issues, from anxiety to family problems to suicidal ideation. 
{The platform assigns each texter to an available counselor; under the assignment mechanism, counselors cannot select which conversations they take, such that longitudinal changes in behavior cannot be accounted for by shifts in counselors' preferences.}
In addition, many texters only contact the service once and the mechanism does not specifically assign returning texters to the same counselor, distinguishing this setting from psychotherapy contexts with sustained counselor-client relationships considered in other work.
One salient guiding force for new counselors in this challenging domain is the training curriculum,
through which counselors develop skills such as active listening and collaborative problem-solving in a 35-hour-long course. 
{Beyond this initial training}, other forces may also act to shape a counselor's behavior through the remainder of their tenure. For instance, as counselors take more conversations, they are exposed to a broader range of texter situations beyond what the curriculum can account for; 
these experiences may further reinforce or alter their linguistic practices and fill inevitable gaps in the training.\footnote{{Each shift is overseen by supervisors who can offer occasional feedback to counselors during their shifts; counselors can also interact with one another and solicit high-level advice. While in this context the level of social feedback is minimal, an interesting line of future work could more explicitly consider the interplay between an individual's behavioral development and such interactions.}} 
\noindent\textbf{Feasibility check: Is there linguistic change?}
We first provide a coarse demonstration that  the  linguistic behavior of counselors does indeed change with experience in a systematic way.
We consider a toy classification task: determining whether a conversation is taken by a \textit{new} counselor who has experienced less than 20 conversations (\lifestage{0}), versus a \textit{tenured} counselor who has taken between 100 and 120 (\maxlifestage), on the basis of the language in the counselor's messages. In particular, we formulate a \textit{paired} prediction problem of distinguishing between two conversations taken by the \textit{same} counselor at \lifestage{0} and \maxlifestage. 
Over a random subset of 10\% of counselors in our data---comprising 
3,075 conversation pairs---we train a logistic regression model using bigrams in counselor messages as features; we perform 10-fold cross-validation and ensure that no counselor spans multiple folds.\footnote{We use logistic regression models with $\ell_1$ loss, tf-idf transforming features, and grid-searching over $C$, the number and the maximum document frequency of bigrams.} 

The model 
gets a cross-validation accuracy of 86\%, underlining that counselors undergo dramatic linguistic changes with experience; the high accuracy also highlights that such changes exhibit \textit{systematic} consistencies across different counselors.
Moreover, an analogous model that considers texters' messages achieves only 57\% accuracy, suggesting that changes in counselor behavior do not merely reflect changes in the texters they interact {with}.\footnote{That this low accuracy nonetheless exceeds the 50\% random baseline suggests that changes in counselor behavior are reflected in aspects of a texter's behavior as well. Indeed, prior work \cite{althoff_large-scale_2016} has suggested that texters can be influenced within a conversation; future work could further model whether the nature of this influence on the texter also varies with counselor experience.}
This linguistic development is also not solely encompassed by population-wide shifts in trivial surface-level attributes---there is no systematic change {in the word-counts of counselors' messages or the lengths of their conversations}.
In the remainder of this work, we will develop a methodology to characterize the systematic nature of this linguistic change.

\section{Measures of Linguistic Diversity}
\label{sec:method}

As discussed in the introduction, certain circumstantial forces can drive counselors to diversify their conversational language with experience, while others can lead to linguistic stagnation. 
In order to operationalize these  intuitions, we first design a general framework aimed at capturing the degree of \textit{linguistic diversity} across conversations and individuals.\footnote{{To encourage its application to other conversational domains, we release an implementation of this framework as part of ConvoKit:} \url{https://convokit.cornell.edu}} 
We then  instantiate this framework in the counseling domain and estimate the relation between the linguistic diversity of counselors and 
their effectiveness in engendering positive outcomes.
Finally, we use the resulting high-level linguistic characterization to analyze the \emph{evolution} of a counselor's behavior over their tenure---the main focus of this work.

\xhdr{Types of diversity}
We can measure linguistic diversity in conversations relative to two reference points.
First, an individual can linguistically vary \textit{across the different conversations} they have; second, their language can deviate from that of \textit{other individuals}. 
We develop two complementary language-model-based measures corresponding to these reference points.

\xhdr{Within-individual diversity} Formally, for a particular individual \inter{I} and life-stage \lifestage{i}, we divide the set of conversations in that stage into two temporally-interleaved subsets, henceforth the \textit{training} and \textit{test} sets. We train a unigram language model \LM{I}{i} over all of the individual's messages in conversations in the training set. For each conversation $c$ in the test set, we can then compute its \textit{within-cross-entropy} \crossent{c}{\LM{I}{i}} to \inter{I}'s language model in \lifestage{i}.  We average within-cross-entropies across all test set conversations 
to quantify \inter{I}'s within-individual diversity during \lifestage{i}.\footnote{For computational efficiency, we account for unseen vocabulary with an approximate form of smoothing by treating unseen words as tokens with frequency 1 in the training set.}

\xhdr{Between-individual diversity} We can analogously define \inter{I}'s between-individual diversity by replacing \inter{I}'s language model at life-stage \lifestage{i} with the language model of another individual \inter{J} at the same life-stage, where \inter{J} is randomly-selected among \inter{I}'s peers. 
In this way, we calculate the \textit{between-cross-entropy} \crossent{c}{\LM{J}{i}} for each conversation $c$ in \inter{I}'s test set of conversations, which we average to quantify \inter{I}'s between-individual diversity during \lifestage{i}.

Cross-entropy is known to be sensitive to the amount of text used to train \LM{I}{i}, as well as to the length of the document $c$ \cite{genzel_entropy_2002}. To mitigate trivial length-based effects and ensure that our measures of diversity afford meaningful comparisons across life-stages and across individuals, we randomly sample $W$ words from each training set to construct the language models, and 
$w$ words from each conversation to compute the  cross-entropies.  For robustness, we take averages over several samples {for each measure, and over} randomly selected peers \inter{J} for between-cross-entropy.

\xhdr{Relative diversity} We note that within- and between-individual diversity are intrinsically related.  If an individual's own conversations are closer to each other than they are from conversations of other individuals we can say that the counselor has a \emph{distinctive voice}.  This would not be the case if the between-individual diversity would be entirely explained by their within-individual diversity.  To capture this intuitive notion we can measure \inter{I}'s linguistic distinctiveness during each \lifestage{i} as the average difference between the within- and between-cross-entropies of each of their conversations in the corresponding test set. 
We interpret high relative diversities as indicative of individuals who have a consistent voice that stands out when compared to others.

\xhdr{Application to the counseling domain}
To apply these methods to analyze mental health counselors, we construct language models from $W\,=\,2,000$ words of each counselor and life-stage, compute cross-entropies with $w\,=\,200$ words per conversation and average over $i\,=\,50$ {random samples}.\footnote{These parameters are chosen to ensure sufficient data was used to characterize a counselor's language.
}  
To account for potential differences in peer groups that may have been
{trained with different curricula,} for between-counselor diversity we consider as \emph{peers} counselors who started taking conversations in the same month.
\xhdr{Relation to counselor effectiveness} 
Before we analyze the dynamics of linguistic diversity, we briefly consider its downstream implications for conversation outcome---to what extent is linguistic diversity positively reflective of a counselor's skill? 
Higher diversity can signal the ability to hold more natural interactions and to better address specific texter situations, which according to literature might translate into positive counseling outcomes \cite{gaume_counselor_2010,atkins_scaling_2014,perez-rosas_analyzing_2018}.

Determining the quality of a counseling conversation is a fundamental difficulty in this domain \cite{tracey_expertise_2014}.
%
%
As one imperfect indicator of quality, we make use of texter responses to a survey given after each conversation, asking the texter whether or not they felt helped by the interaction.
In our data, 26\% of conversations received a rating, out of which 87\% were rated as helpful. 
We consider the proportion of positive ratings a counselor received in a particular life-stage as an indication of their effectiveness.\footnote{Here, we ignore conversations where no response was received and consider only counselors in life-stages where they received at least four ratings. 
}
To gauge the implications of a counselor's linguistic diversity, we compare the effectiveness of the most and least diverse counselors, in terms of both diversity measures.
In the following discussion, we focus on experienced counselors, in the life-stage comprising their  80th to 120th conversations,\footnote{Our results are similar, with some loss of significance, for alternate experience levels.} and compare the top and bottom third of counselors for each measure.

We find that both within- and between-individual diversities are positive signals of effectiveness---for instance, the average effectiveness of the most within-individual diverse counselors is 89\% while the effectiveness of the least within-individual diverse counselors is 86\%
(Mann Whitney U test $p < 0.001$). 

We note that the observed correspondences between diversity and effectiveness have a potentially trivial explanation: a counselor may have an easier time varying their language across {a series of texters with less complicated issues who may more readily give good ratings}, independent of the counselor's actual skill. 
To discount this somewhat circular
explanation, we can estimate the counselor's diversity in an earlier life-stage (40th to 80th conversations), and relate it with their effectiveness in the \textit{subsequent} life-stage (80th to 120th conversations).
The above differences still hold with statistical significance in this new setting, showing that diversity is non-trivially informative of a counselor's \emph{future} effectiveness.\footnote{Developing a {model that can estimate} the future effectiveness of a counselor constitutes an interesting avenue for future work.  We note that this task would be distinct from that of predicting the effectiveness of a presently-observed conversation, as was addressed in prior work \cite{atkins_scaling_2014,althoff_large-scale_2016,perez-rosas_analyzing_2018}.}
\begin{figure*}[t]
\centering
\includegraphics[width=0.8\textwidth]{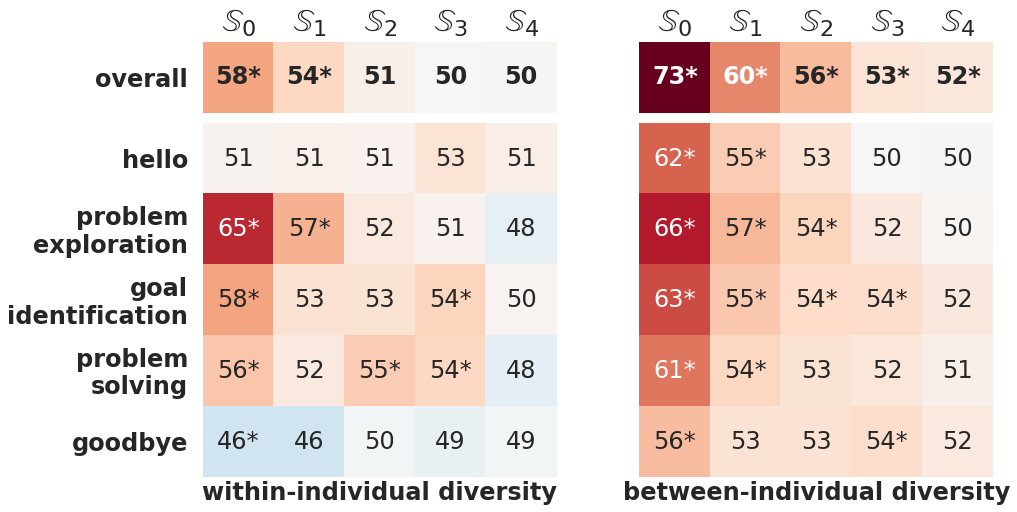}
\caption{
Temporal diversification trends across different counselor life-stages (left to right). Each cell shows the percentage of counselors that increase in diversity from that 
life-stage to their tenured life-stage \maxlifestage; *'s indicate statistical significance (binomial test $p<0.05$, comparing to 50\% by chance).  The topmost row shows the temporal trend across the entire conversation, and the subsequent rows separate this trend by conversational component.  
}
\label{fig:increase_heatmap}
\end{figure*}

\section{Analyzing Counselor Diversification}
\label{sec:divergence}

We are now equipped to analyze how linguistic diversity develops as counselors gain experience.

\xhdrq{Do counselors diversify}
We first quantify the extent to which any development occurs. To this ends, we compute the percentage of counselors that increased their diversity from their initial conversations during \lifestage{0} to their tenured life-stage \maxlifestage.
We find that, overall, counselors tend to diversify as they gain experience: 58\% increase in within-individual diversity---indicating that they grow to vary their language to a greater extent between different conversations they take---, and 73\% increase in between-individual diversity---indicating that they become more linguistically distant from their similarly-experienced peers.   
Taking these measures together, we also find that most counselors increase in relative diversity (72\%), becoming linguistically further away from their peers than from themselves.
Overall, these trends suggest an increasing inclination for counselors to move away from an initial linguistic rigidity, eventually finding their own distinctive {voice}.\footnote{{We note that the time it takes for the counselors to reach their tenured life-stage is only mildly correlated to each diversity measure ($\rho < 0.2$ for each).}}

\xhdrq{When do counselors diversify}
Is this observed diversification sustained over the counselors' tenures, 
or concentrated in an initial change?
Answering this question is important for understanding, and potentially assisting, the {developmental trajectory} of mental health counselors \cite{tracey_expertise_2014}.
As before, we consider as a reference the language of tenured counselors during \maxlifestage, and show in Figure \ref{fig:increase_heatmap} (top row) the percentage of counselors that increase in diversity between each of their earlier life-stages and \maxlifestage. We observe that both types of diversity increase over an extended span of counselors' tenures, but gradually level off to match the diversity of \maxlifestage (i.e., percentages approach 50\%), suggesting that linguistic development {plateaus} with experience.

\begin{figure*}[t]
\centering
\includegraphics[width=1\textwidth]{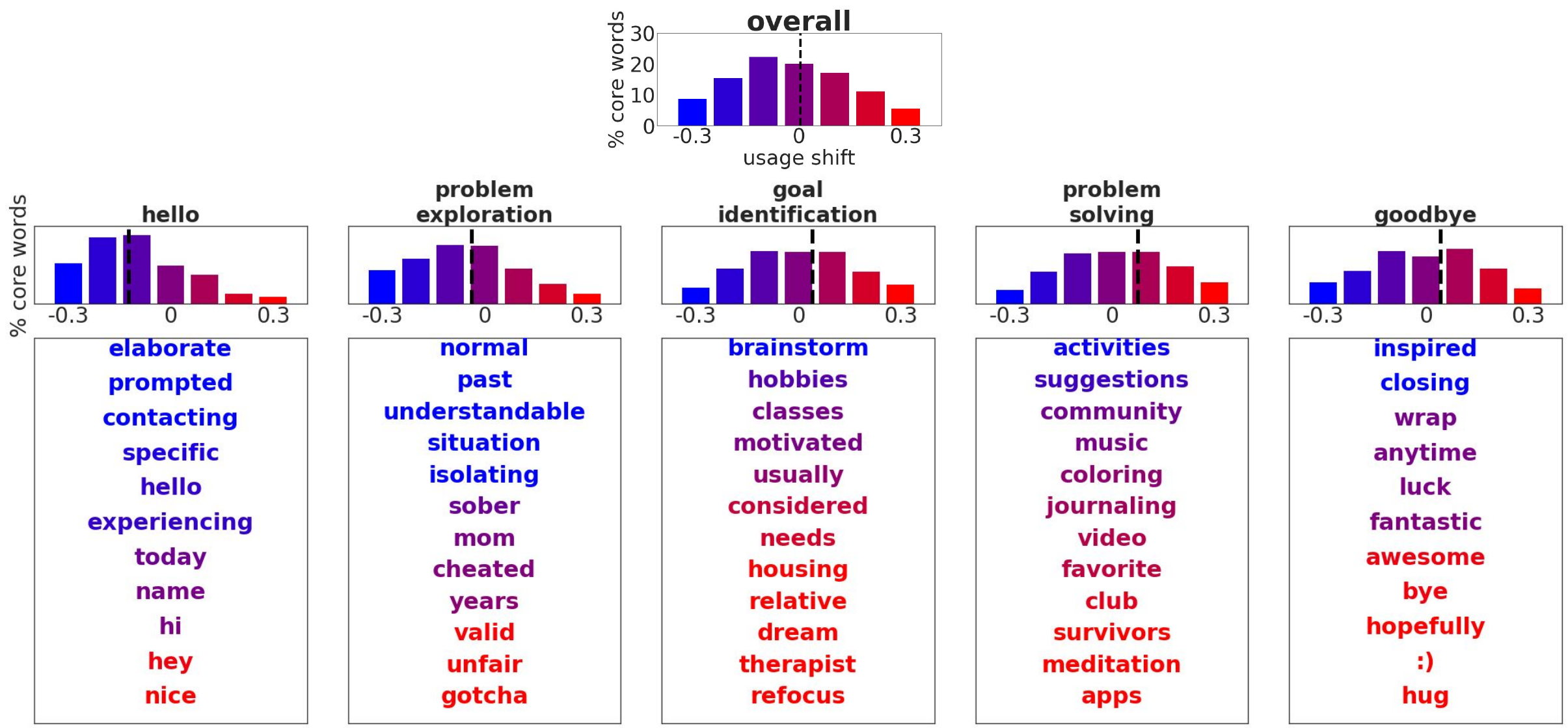}
\caption{
	\textbf{Top}: the overall distribution of usage shifts of the words 
	{at the core of the}
	counselor vocabulary (i.e., those used by at least 20\% of counselors). 
	\textbf{Bottom}: 
	{usage shift distributions per}
	subset of core words characteristic to each conversational component, along with examples.
	Words are ordered by, and colored according to their usage shift, such that \textcolor{blue}{blue} words tend to be used more by counselors in their earlier conversations while \textcolor{red}{red} words tend to be used more by counselors with more experience.
	Dashed lines indicate the respective medians.
}
\label{fig:usage_shift_hist_segs}
\end{figure*}

\xhdrq{What diversifies}
Conversations are not uniform---different aspects of the interaction may engender differing evolution dynamics. In this particular domain, conversations follow a well-defined structure that is taught as a central element of the training curriculum, proceeding through functionally distinctive segments---henceforth \emph{components}---where counselors (1) build initial rapport with the texter (\firstseg), (2) explore the texter's problems (\secondseg), (3) identify the texter's goals and past attempts to cope (\thirdseg), (4) collaboratively explore steps towards achieving those goals (\fourthseg), and finally (5) close the conversation (\fifthseg). As a rough proxy to capture these functional components, we divide a conversation into fifths, as in prior work \cite{althoff_large-scale_2016}.\footnote{For this analysis, we only consider conversations with at least 20 counselor messages to ensure meaningful conversation segments. While we found that this coarse segmentation was satisfactory for our analyses, future work could consider more involved approaches to segmenting conversations.} 
Characteristic 
words for each component are shown in Figure \ref{fig:usage_shift_hist_segs}.
To chart linguistic development with respect to our measures across functionally-different aspects of the conversation, we compute each measure separately over the messages in each  fifth of the conversation, and repeat our temporal analysis per component (Figure \ref{fig:increase_heatmap}, bottom rows).

We observe that development in within-individual diversity is largely concentrated in situation-specific 
segments of the conversation (\secondseg, \thirdseg, \fourthseg); in these segments the development is sustained throughout counselor tenures.  This could suggest that as counselors encounter more diverse situations, they become linguistically better equipped  to approach them in a specialized fashion.  In contrast, in conversational components which have functions that generalize across texter problems there is no systematic increase in diversity (\firstseg), and there is even a slight decrease (\fifthseg)---perhaps indicating that counselors develop routines to close the conversation. 

On the other hand, between-individual diversity exhibits a comparatively more uniform increase across each component. This could suggest that counselors adopt individually distinguishing linguistic styles throughout the conversation.

In the next section we provide further support for these intuitions through a finer-grained analysis of vocabulary changes.

{\section{Development of Counselor Vocabulary}}
\label{sec:lexical}

Thus far, we have tracked the evolution of counselors according to high-level characterizations of their linguistic diversification. We now analyze this evolution at a complementary granularity: tracking changes in counselors' use of different \textit{words} over the course of their tenure, as tangible and interpretable indicators of linguistic development.
This analysis offers concrete examples of lexical changes that reflect the intuitions gleaned from our preceding examination.

A counselor may grow to use some words more often, perhaps reflecting continued exposure to particular texter situations or the development of a personal style. 
Such adoptions could enrich and diversify a counselor's vocabulary, and perhaps distinguish them from others; they could also mark aspects that converge to a common counselor language, counterbalancing the diversification process.
Alternatively, some words may fall out of favor, getting used less frequently over time. 
This possibility is especially pertinent given the initial training process, since the
  curriculum illustrates many of the counseling practices by way of presenting examples of messages for counselors to build on. 
{These serve as a linguistic reference point for starting counselors; linguistic divergence could then arise as more experienced counselors start to move beyond this reference point.}
{To examine such changes in counselor vocabulary,} 
we quantify the population-wide \textit{usage shift} of a word 
as the log-ratio between the proportion of conversations in \maxlifestage in which counselors used that word, and the proportion of conversations in \lifestage{0} in which counselors used that word.\footnote{To avoid spurious effects akin to Simpson's paradox, we only consider conversations taken by counselors with at least 120 conversations, such that each counselor contributes equally to both the \maxlifestage and \lifestage{0} components of the measure.}
Thus, a positive shift indicates that the word's usage increases with tenure, while a negative shift means that its usage declines after the counselor's first
conversations.
Figure \ref{fig:usage_shift_hist_segs} 
 {(top)} shows a histogram of usage shifts {for the \textit{core} of the counselors' vocabulary, i.e., the 3,461 words used by at least 20\% of the long-serving counselors considered in the preceding analyses.}

\xhdr{The interplay of lexical shifts and function}
Our prior analyses revealed that diversification patterns vary by functionally-different conversational components. To tie usage shifts to these observed contrasts, we examine the usage shifts of words characteristic to each component. Concretely, we consider a word to be characteristic of a component if the proportion of messages containing it in the corresponding fifth of the conversation is greater than the proportion of messages with the word overall (in particular, the log-ratio of the within-component frequency of the word to its overall frequency is at least 0.2). 
Figure \ref{fig:usage_shift_hist_segs} {(bottom)} shows usage shifts over words in each component, together with example words sorted by their shift. 

Across each component, we see that counselors shift away from using words which are relatively general and more formal in tone (negative shift, in blue). 
Many of these words {reflect} counseling practices as presented in the training material, {echoing the language used in training to exemplify these practices.}
For instance, during the first component (\firstseg), counselors are told to build rapport and bootstrap the {exploration of the texter's concerns} by having texters \egwordold{elaborate} on what \egwordold{prompted} them to \egwordold{contact} the platform; as counselors 
{proceed in this exploration process} (\secondseg) they are taught to reflect on and validate the texter's concerns as \egwordold{understandable} and \egwordold{normal}. 

That counselors eventually use such words less often does not necessarily mean that 
{the practices underlying them} are abandoned with experience; rather, counselors may gravitate towards new words that accomplish the same purposes (\egwordnew{valid}, \egwordnew{unfair}).
Words used in the first two components are especially prone to {decreasing in usage} with experience. 
This suggests that while incoming counselors use the training material as a linguistic guide for initiating conversations, they {tend to eventually} shift towards more informal and less {standardized} {language, 
moving away from using words in the core of the counselors' vocabulary.}\footnote{{We note that our usage shift methodology only captures population-wide changes in how counselors use a subset of core words, so it cannot characterize changes outside this general vocabulary. Future work could examine individual-level lexical changes (such as the development of uniquely personal catchphrases).}} 

Conversely, many words which counselors adopt later in their tenure
(positive shift, in red) are more specialized, especially in the \thirdseg and \fourthseg components.
Intuitively, counselors accumulate domain knowledge with experience, {allowing them to better differentiate between cases}, especially in terms of potential coping mechanisms (e.g., \egwordnew{therapist}, \egwordnew{meditation}, \egwordnew{apps}). 
{
This type of specialization suggests that much of the broadening in within-individual diversity we observed in these conversational segments (Figure~\ref{fig:increase_heatmap}, left) may be
functional in nature.}

Other words which increase in usage with experience are characterized by a relatively colloquial tone (\egwordnew{hey}, \egwordnew{:)}, \egwordnew{gotcha}), pointing to a diversification {process} that is 
{social rather than functional.}
This linguistic relaxation may {trace} the observed trend towards higher between-individual diversity (Figure~\ref{fig:increase_heatmap}, right) as counselors become more at ease in their task. {Shifts towards such colloquial language are especially prominent in the last component (\fifthseg), potentially as counselors develop individual routines for closing conversations, and} echoing the decrease in within-individual diversity observed in that component (Figure~\ref{fig:increase_heatmap}, left). 


Overall, these observations complement our prior discussion about the rate at which counselors diversify their language, suggesting functional and social mechanics that may be in play.

\vspace{0.15in}
\section{Discussion and Future Work}
\label{sec:discussion}

Having good conversations is challenging and hard to teach, especially in domains where the stakes are high and where feedback is scarce.  
Computational tools could help us understand how people acquire conversational skills, and may eventually assist in their development.  In this work we provide an initial 
and limited case study in a highly consequential domain, showing that we can track {diversification in the linguistic practices of} mental health counselors.

Our approach is necessarily limited in scope.  Future work could adapt our framework to examine more complex {forms} of conversational development. In particular, we observed that counselors diversify linguistically, perhaps signaling a beneficial increase in flexibility that enables them to better address the specific concerns of a texter. Future work could more directly model how counselors {respond} to texter behaviors, hence gauging the extent to which counselors evolve in their \textit{interactional} practices.  Furthermore, while our approach captures linguistic changes in aggregate, {a complementary line of work could} explore the trajectories of individuals, and probe the factors determining whether particular individuals diversify or stagnate.

{
The framework we have started to design could eventually help platforms provide counselors with personalized feedback on their development. It could assist in identifying counselors who acclimatize quickly and those that require more guidance.
Understanding how different components in a conversation change with experience could also inform the 
 particular aspects to focus their training on.
In order to derive such prescriptive recommendations, further work is needed to causally connect our signals of linguistic experience with actual {expertise} and {skill}, as reflected in conversational outcomes. 
For instance, increased diversification may point to flexibility and specialization that are beneficial, but might also signal detrimental deviations from good counseling practices. 
Other approaches, such as qualitative labeling by domain experts, could examine whether such changes in language use also result in better conversations. 
}

Our methodology could  also be extended to examine other conversational contexts such as academic advising or business interactions, where individuals are expected to learn from experience. Such domains may contain crucial differences that motivate extensions to our framework; for example, feedback might be more readily available and conversational partners may recur, both of which can interact with experience to further shape linguistic development.

{\small
\xhdr{Acknowledgments}
{Thanks to Crisis Text Line for making this work possible. We are grateful to the anonymous reviewers for their thoughtful comments and suggestions, and to Jonathan P. Chang, Liye Fu, Sendhil Mullainathan and Andrew Wang for helpful conversations. 
This research has been supported in part by  NSF CAREER Award IIS1750615, NSF Grant SES-1741441, and a Microsoft Research PhD Fellowship. The collaboration with Crisis Text Line was supported by the Robert Wood Johnson Foundation; the views expressed here do not necessarily reflect the views of the foundation.}
}

\bibliography{CTL-acl-autoupdate-JZ,CTL-acl-autoupdate-C}
\bibliographystyle{acl_natbib}

\end{document}